\documentclass[letterpaper, 10 pt, conference]{ieeeconf}  

\IEEEoverridecommandlockouts                              

\usepackage[T1]{fontenc}
\usepackage{blindtext}
\usepackage{graphicx}
\usepackage[table]{xcolor}

\usepackage{amsmath}
\usepackage{amssymb}

\usepackage{graphicx}
\usepackage{placeins}

\usepackage{diagbox} 
\usepackage{subfig}
\usepackage{gensymb}
\usepackage{textcomp}
\usepackage{hyperref}
\usepackage{amsmath}

\usepackage{multirow}
\usepackage{balance}

\usepackage{amsmath}

\DeclareMathOperator{\atantwo}{atan2}
\usepackage{mathtools}

\usepackage[colorinlistoftodos]{todonotes}

\usepackage{lipsum}

\newcommand\copyrighttext{%
  \footnotesize \textcopyright 2021 IEEE. Personal use of this material is permitted.
  Permission from IEEE must be obtained for all other uses, in any current or future
  media, including reprinting/republishing this material for advertising or promotional
  purposes, creating new collective works, for resale or redistribution to servers or
  lists, or reuse of any copyrighted component of this work in other works.
  DOI: \href{http://dx.doi.org/10.1109/LRA.2021.3136863}{10.1109/LRA.2021.3136863}}
\newcommand\copyrightnotice{%
\begin{tikzpicture}[remember picture,overlay]
\node[anchor=south,yshift=10pt] at (current page.south) {\fbox{\parbox{\dimexpr\textwidth-\fboxsep-\fboxrule\relax}{\copyrighttext}}};
\end{tikzpicture}%
}

\title{\LARGE \bf
MinkLoc3D-SI: 3D LiDAR place recognition with sparse convolutions, spherical coordinates, and intensity
}

\author{Kamil \.Zywanowski*$^{1}$, Adam Banaszczyk*$^{1}$, Micha\l{} R. Nowicki$^{1}$, and Jacek Komorowski$^{2}$
\thanks{* Equal contribution}
\thanks{$^{1}$ The authors are with the Institute of Robotics and Machine Intelligence,
Faculty of Control, Robotics, and Electrical Engineering,
        Poznan University of Technology, Poznan, Poland
        {\tt\small michal.nowicki@put.poznan.pl}}%
\thanks{$^{2}$  Jacek   Komorowski   is   with the  Faculty   of   Electronics   and   Information   Technology,   Warsaw   University   of   Technology,   Warsaw,   Poland
        {\tt\small jacek.komorowski@pw.edu.pl}}%
\thanks{M. R. Nowicki is supported by the Foundation for Polish Science (FNP)}
\thanks{This research was supported by Nvidia Hardware Grant programme with a single Nvidia A100}
}

\begin{document}

\maketitle

\copyrightnotice

\thispagestyle{empty}
\pagestyle{empty}

\begin{abstract}
The 3D LiDAR place recognition aims to estimate a coarse localization in a previously seen environment based on a single scan from a rotating 3D LiDAR sensor.
The existing solutions to this problem include hand-crafted point cloud descriptors (e.g., ScanContext, M2DP, LiDAR IRIS) and deep learning-based solutions (e.g., PointNetVLAD, PCAN, LPD-Net, DAGC, MinkLoc3D), which are often only evaluated on accumulated 2D scans from the Oxford RobotCar dataset.
We introduce MinkLoc3D-SI, a sparse convolution-based solution that utilizes spherical coordinates of 3D points and processes the intensity of 3D LiDAR measurements, improving the performance when a single 3D LiDAR scan is used.
Our method integrates the improvements typical for hand-crafted descriptors (like ScanContext) with the most efficient 3D sparse convolutions (MinkLoc3D).
Our experiments show improved results on single scans from 3D LiDARs (USyd Campus dataset) and great generalization ability (KITTI dataset). 
Using intensity information on accumulated 2D scans (RobotCar Intensity dataset) improves the performance, even though spherical representation doesn't produce a noticeable improvement.
As a result, MinkLoc3D-SI is suited for single scans obtained from a 3D LiDAR, making it applicable in autonomous vehicles.
\end{abstract}

\section{Introduction}

Place recognition modules, determining if the sensor revisits a previously observed location, are vital whenever a long-time autonomous operation is required, i.e., for simultaneous localization and mapping (SLAM).
In SLAM, the ability to relocalize can reduce the accumulated localization drift while correcting the past trajectory to build a consistent map of the environment.
Among place recognition applications, robust place recognition for autonomous cars is a commonly tackled problem. 
City-wide localization in dynamic environments with moving objects, changing weather conditions, and seasonal changes requires robust methods capable of capturing low-level and high-level features from raw sensory data. 

\begin{figure}[t]
    \centering
    \includegraphics[width=\columnwidth]{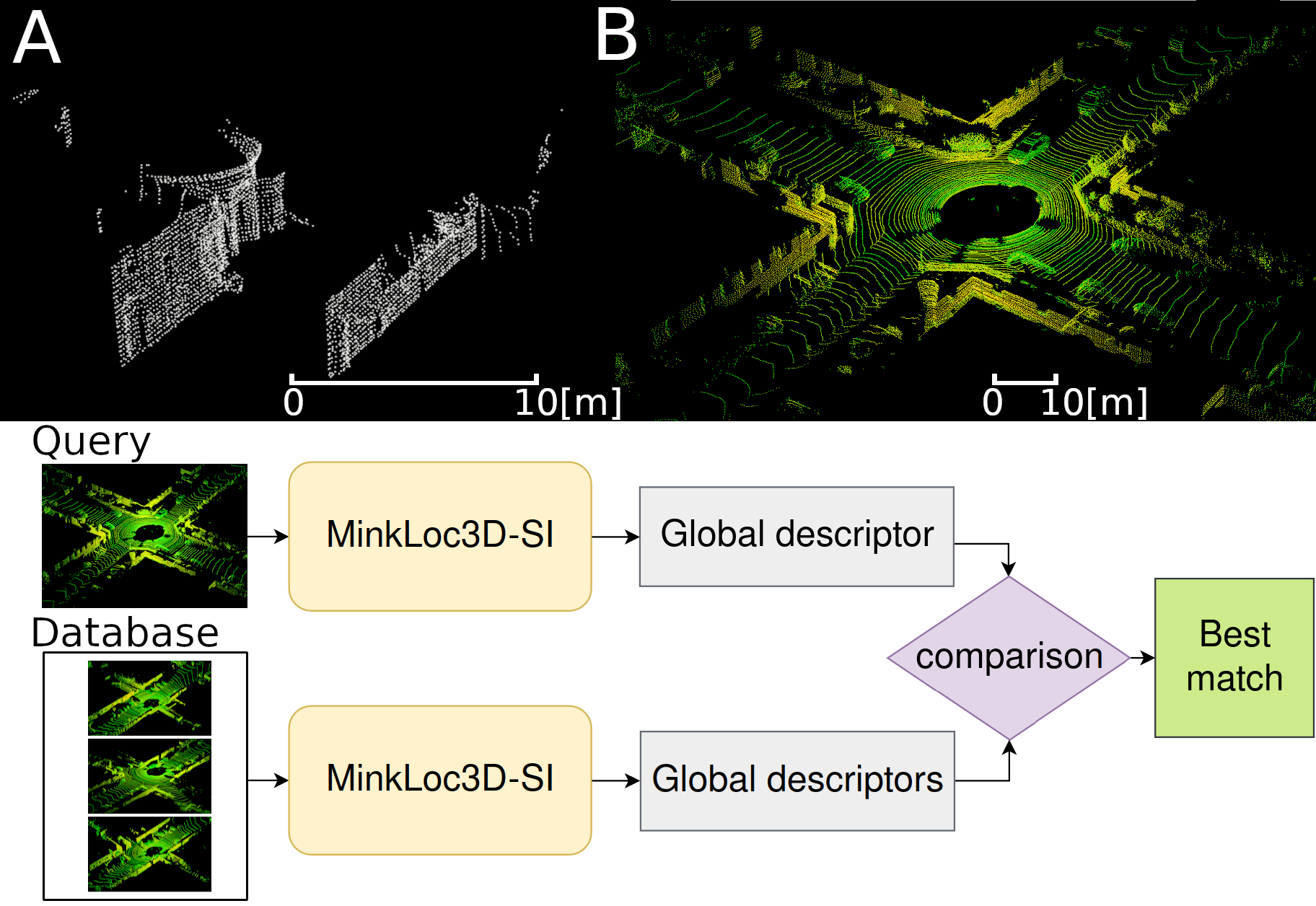}
    \caption{The LiDAR place recognition solutions are most commonly evaluated on combined 2D scans containing 4096 equally distributed points from the Oxford RobotCar dataset (A). In contrast, actual 3D scans have a single point of observation, contain more points, and include intensity as in the KITTI dataset (B). The proposed MinkLoc3D-SI targets the irregular distributions of points in the 3D LiDAR scans while utilizing intensity to generate descriptors that are compared to determine if the query location matches one of the locations in the database.}
    \label{fig:catchy}
\end{figure}

Designing an efficient learning-based 3D LiDAR place recognition system is still an open problem. 
A key challenge is to find the best 3D data representation that can be efficiently processed using neural networks to extract meaningful features for robust place recognition.
As a community, we have already explored image-like~\cite{icarcv}, voxel-based~\cite{voxelnet}, bird's-eye view approaches~\cite{pointpillars,scanContext,scancontext++}, or unordered point sets~\cite{pointnetvlad,NDTtransformer} representations.
More recently, we also see a rise in sparse convolution-based approaches~\cite{minkloc3d,minklocplusplus} and attention-based modules~\cite{attdlnet} that might be combined together~\cite{transloc3d} as well.
Many of these methods are trained, evaluated, and compared on the Oxford RobotCar dataset~\cite{pointnetvlad,minkloc3d,pcan,lpdnet,dh3d,dagc,epcnet,soenet}, which was gathered by concatenating multiple 2D LiDAR scans covering a 20~m distance and subsampling to 4096 points~\cite{pointnetvlad}.
In contrast, a single scan from a modern 3D LiDAR covers a much larger area (even up to 160-200 meters in diameter) and has a greater number of points (up to 260k points for Ouster OS1-128), as presented in Fig~\ref{fig:catchy}. 

We propose a new 3D LiDAR place recognition system called MinkLoc3D-SI, extending the concept of sparse convolution-based MinkLoc3D~\cite{minkloc3d} to improve performance on scans from 3D LiDARs.
The proposed MinkLoc3D-SI uses spherical coordinates of 3D points and utilizes the intensity value in 3D LiDAR scans.
We evaluate our approach on the USyd Campus~\cite{usyd}, Oxford RobotCar~\cite{oxford}, and KITTI~\cite{kitti} datasets.

The main contributions of our work are:
\begin{itemize}
    \item The first 3D sparse convolution-based place recognition system, MinkLoc3D-SI, utilizing intensity and spherical point representation suited for place recognition based on a single scan from 3D LiDAR.
    \item A new place recognition dataset utilizing Velodyne VLP-16 based on the USyd Campus dataset.
    \item A modified Oxford RobotCar Intensity dataset including intensity for each 3D point.
\end{itemize}
All the code and datasets enabling the replication of results presented in the paper are publicly available\footnote{\url{https://github.com/KamilZywanowski/MinkLoc3D-SI}}.


\section{Related work}
\subsection{3D LiDAR data representation for deep learning}

The community explored different options of 3D point cloud representation for deep learning.
We started with 2D range image representations processed by 2D convolutions~\cite{icarcv}, being the closest representation to those typically used for RGB images.
Later approaches focused on the volumetric representations~\cite{voxelnet} that capture the structure of the 3D point cloud.
The problem is that their computational complexity usually grows cubically with the number of voxels. 
In turn, the community explored the possibility of reducing dimensionality to increase efficiency, i.e., by introducing a 2.5D bird's-eye view point pillars~\cite{pointpillars} representation, suitable for autonomous cars.

An alternative approach assumes processing raw point clouds, e.g., using a PointNet-like~\cite{pointnet,pointnet++} architecture.
In the PointNet architecture, to enforce invariance to permutations of the input point cloud, a large part of the processing is done separately for each point.
Thus, this design is not well-suited to extract informative global features.

The most recent approaches use 3D convolutional architectures and a sparse volumetric representation of a point cloud. These methods are efficient due to optimized implementations which use hashing-based addressing techniques to quickly perform convolutions on sparse data~\cite{minkowski,sparse2}.
As a result, this representation is gaining popularity with successful applications in many areas, such as semantic segmentation of 3D LiDAR scans~\cite{spvnet}.

\subsection{3D LiDAR place recognition}

Existing 3D LiDAR place recognition solutions are closely related to developments in data representation.
The range image-based representation of LiDAR scans is used in~\cite{icarcv} to achieve localization robust to changing weather conditions.
More recently, X. Chen \textit{et al.} presented OverLapNet~\cite{overlapnet}, which combines image-based representation with normal and semantic layers for metric localization. 
In AttDLNet~\cite{attdlnet}, authors combine the proxy representation of range images with attention modules to prove that attention is a critical component in the descriptor learning.

The first notable example of a method operating on a raw 3D point cloud is PointNetVLAD~\cite{pointnetvlad}. 
It combines PointNet~\cite{pointnet} local feature extraction with NetVLAD~\cite{netvlad} to aggregate local features into a global place descriptor.
Unfortunately, the method suffers from PointNet weakness to capture high-level features.
Therefore, many solutions like~\cite{pcan,lpdnet,dagc} focuse on the data representation problem leaving the NetVLAD part intact.
PCAN~\cite{pcan} improves PointNet by estimating the significance of each point. DAGC~\cite{dagc} uses graph CNN to combine information at multiple scales.
LPD-NET~\cite{lpdnet} computes hand-crafted features, which are later processed using a pipeline similar to PointNet architecture.
The most recent method, NDT-Transformer~\cite{NDTtransformer}, combines local distribution of raw points with Normal Distributions Transform (NDT). Combined features are processed by the Transformer module and aggregated using NetVLAD to achieve state-of-the-art performance.


SegMap~\cite{segmap} presents another approach for combining individual scan features, based on segments extracted from a 3D point cloud.
This approach is extended in~\cite{janps} by combining SegMap with intensity information to improve descriptor discriminativity. 
Locus~\cite{locus} analyzes topological relationships between segments in a single scan and temporal correspondences between segments in a sequence of scans.

A popular approach is to represent the scan as the bird's-eye view (BEV) image.
Scan Context~\cite{scanContext} is a hand-crafted descriptor efficiently computed from a BEV scan representation in polar coordinates.
Scan Context sparked a new family of BEV approaches, including a trained descriptor called DiSCO~\cite{disco}, Scan Context augmented with intensity information~\cite{intensitySC}, or semantic-based extension called SSC~\cite{li2021ssc}.
Recently presented Scan Context++~\cite{scancontext++} extends Scan Context, by providing metric localization on top of the existing topological localization.
In BVMatch~\cite{bvmatch}, BEV is combined with Log-Gabor filters to detect keypoints later used in a typical topological localization with a bag-of-visual-words (BoVW) approach for topological and metric localization.
 
The availability of the efficient 3D sparse convolution library sparked an interest in using sparse volumetric representation for place recognition purposes~\cite{minkloc3d,minklocplusplus,transloc3d}. 
The first method, MinkLoc3D~\cite{minkloc3d}, surpasses previous methods by a significant margin when evaluated on the Oxford RobotCar dataset, proving that the data representation is a critical component of a 3D LiDAR place recognition method.
MinkLoc++~\cite{minklocplusplus} further improves MinkLoc3D with a channel attention mechanism ECA~\cite{ecanet} while fusing 3D LiDAR scans and camera images. 
Similar approach is used in TransLoc3D~\cite{transloc3d}, which combines sparse convolutions, adaptive receptive field module (ARFM) based on ECA, and Transformer module.

Our work combines the idea presented in Scan Context~\cite{scanContext} to use non-Cartesian point representation well suited for 3D LiDAR scans with efficient sparse 3D convolutional architecture while utilizing intensity information available for each 3D point.

\section{Proposed solution}

\subsection{MinkLoc3D}

\begin{figure}[htbp!]
    \centering
    \includegraphics[width=\columnwidth]{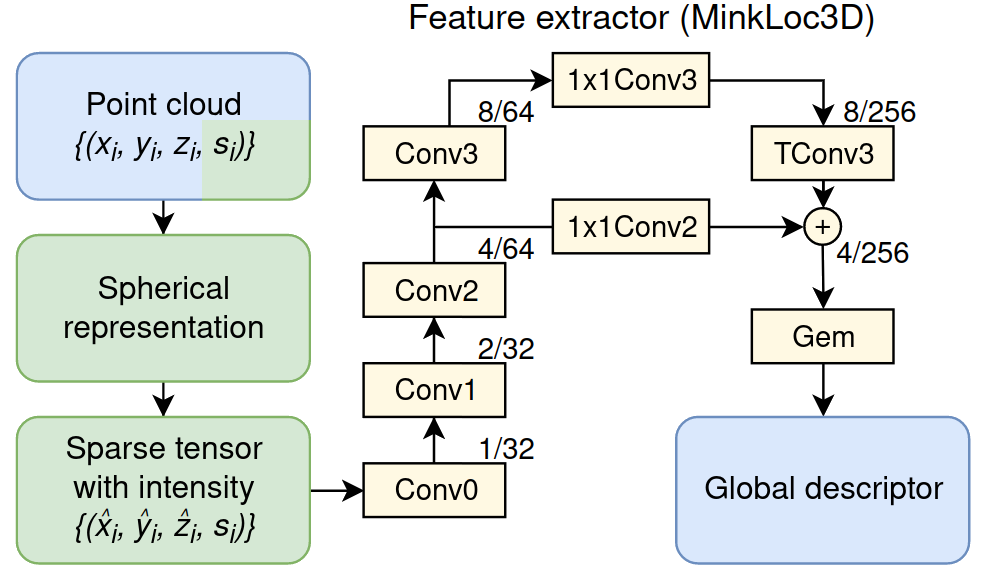}
    \caption{Overall processing steps of the MinkLoc3D-SI architecture. The modifications proposed to the MinkLoc3D model, marked by green color, include spherical representation (MinkLoc3D-S) and intensity usage (MinkLoc-I) combined into MinkLoc3D-SI. Presented numbers (e.g., 1/32) indicate a stride and number of channels of a feature map produced by each block in the local feature extractor.}
    \label{fig:procsteps}
\end{figure}

MinkLoc3D~\cite{minkloc3d} network architecture is based on a local feature extraction network utilizing sparse convolutions and generalized-mean (GeM) pooling layer~\cite{gem}. 
In this approach, presented in Fig.~\ref{fig:procsteps}, the input 3D LiDAR point cloud (${x_i, y_i, z_i}$) is quantized into a finite number of cuboids with a single 3D point located in each cuboid.
The processed information is then used to create a single sparse tensor of quantized points $\mathbf{\hat{C}}$ and an associated feature input $\mathbf{F}$~\cite{minkowski}:
\begin{equation}
    \mathbf{\hat{C}} = \begin{bmatrix} \hat{x}_1 & \hat{y}_1 & \hat{z}_1 \\
    & \vdots & \\
    \hat{x}_N & \hat{y}_N & \hat{z}_N,
    \end{bmatrix}, \mathbf{F} = \begin{bmatrix} {f}_1^T \\
   \vdots\\
    {f}_N^T,
    \end{bmatrix}.
\end{equation}

In the case of MinkLoc3D, it is a purely geometrical approach with each associated feature equal to one $\left({f}_i=\begin{bmatrix}1\end{bmatrix}\right)$ indicating the existence of a 3D point.
The local feature extraction part of the network is built using Feature Pyramid Network (FPN)~\cite{fpn} design pattern. 
The bottom-up part contains four convolutional blocks (Conv0..3), producing sparse 3D feature maps with decreasing spatial resolution and increasing receptive field.
The top-down part contains a transposed convolution (TConv3) generating an upsampled feature map. 
The upsampled feature map is concatenated with features from the corresponding layer of the bottom-up pass using a lateral connection ($1\times1\mathrm{Conv}$). Such architecture produces a feature map with a large receptive field and high spatial resolution~\cite{minkloc3d}.
As a result, the feature extraction network generates a sparse feature map ($h_j^1, h_j^2, ..., h_j^{256}$) for $j$-th non-zero element of the sparse local feature map.
Feature map produced by a local feature extraction part of the network is pooled with generalized-mean (GeM) pooling \cite{gem} to produce a global point cloud descriptor $\textbf{g}$, with its k-th component ${g}^k$ equal to:
\begin{equation}
{g}^k = \left(\frac{1}{M} \sum_{j=1}^{M} \left({h}_j^k\right)^p \right)^{\frac{1}{p}},
\end{equation}
where $M$ is the number of non-zero elements in the sparse local feature map and $p$ is a learnable parameter of GeM pooling that determines a smooth transition from the global max pooling to the global average pooling.

The network is trained with a triplet margin loss~\cite{triplet2,triplet}.
The goal of the triplet loss is to minimize the distance between descriptors of point clouds showing the same location (a reference point cloud descriptor $a$ and its positive match descriptor $p$) and maximize the distance between descriptors of points clouds representing different locations ($a$ and its negative match descriptor $n$):
\begin{equation}
    L = \sum_i \max( d(a_i, p_i) - d(a_i, n_i) + m, 0),
\end{equation}
where $d(x,y)$ is the Euclidean distance between global descriptors $x$ and $y$, and $m$ is a chosen margin parameter.
We use a hard negative mining strategy to ensure that only triplets with non-zero loss $L$ (active triplets) are selected for batch processing.
We utilize data augmentation: random jitter, random removal of points, random translations, random flip, etc.
The network is trained with Adam optimizer with learning rate $l_r=1e^{-3}$ and decay $d=1e^{-3}$.
We used a single Nvidia RTX 3080 card for training, which lasted approximately 4 hours due to efficient hard-negative triplet mining.
More information about the 3D sparse convolutions can be found in~\cite{minkowski} while the in-detail presentation of MinkLoc3D is given in~\cite{minkloc3d}.

\subsection{MinkLoc3D-SI} 

3D point clouds captured by the 3D LiDARs have a varying density with more points closer to the scanner's origin and usually a smaller range of values in the elevation direction. 
Therefore, a regular grid of Cartesian coordinates is not well-suited for 3D points further away from the scanner. The distances between points naturally increase, making it harder to extract high-level features.


We propose MinkLoc3D-S utilizing a spherical representation of 3D coordinates of points, which is a natural representation of 3D LiDAR's measurements.  
MinkLoc3D-S performs sparse convolutions using this alternative representation.
In the proposed approach, each 3D point $(x,y,z)$ with elevation represented by the $z$ component is converted into a corresponding spherical representation $(r, \theta, \phi)$:
\begin{align}
    r &= \sqrt{x^2 + y^2 + z^2}, \\\theta &= \atantwo(y,x), \\\phi &= \atantwo(z, \sqrt{x^2 + y^2}),
\end{align}
where $r$ is the distance between the 3D point and the scanner, $\theta$ is the horizontal scanning angle, while $\phi$ is the vertical scanning angle.
As a result, the area of the quantization cuboid increases for 3D points further away from the 3D LiDAR.

The original MinkLoc3D utilizes only the geometry of 3D points to perform place recognition.
But as stated in~\cite{lidarintensity,lidarintensity2}, the intensity of the returned signal for the 3D LiDARs is valuable and
can be used to construct the local place recognition descriptor.
Therefore, we propose to include the information about the LiDAR intensity in the sparse convolutions of MinkLoc3D-I.

The intensity $s_i$ for the $i$-th 3D point is filled in the feature map part of the processing ${f}_i = \begin{bmatrix} s_i \end{bmatrix}$, which does not increase the dimensionality of the convolutions.
The raw intensity values are normalized to $0-1$ interval.
When multiple points fall into the same quantization cuboid, we randomly choose one of the intensity values from the cuboid during training to ensure proper robustness. Still, we average the intensity values during inference to achieve repeatable results.


%
MinkLoc3D-SI combines both improvements: spherical representation and usage of measurement intensities.

\section{Datasets}

\subsection{USyd Campus}

The USyd Campus Dataset (USyd)~\cite{usyd} contains recordings from a buggy-like car registered over 50 weeks in varying weather conditions.
The used sensory setup consists of a Velodyne VLP-16 LiDAR, cameras, and GPS that serve as ground truth in our application.

\begin{figure}[h!]
    \centering
    \includegraphics[width=0.8\columnwidth]{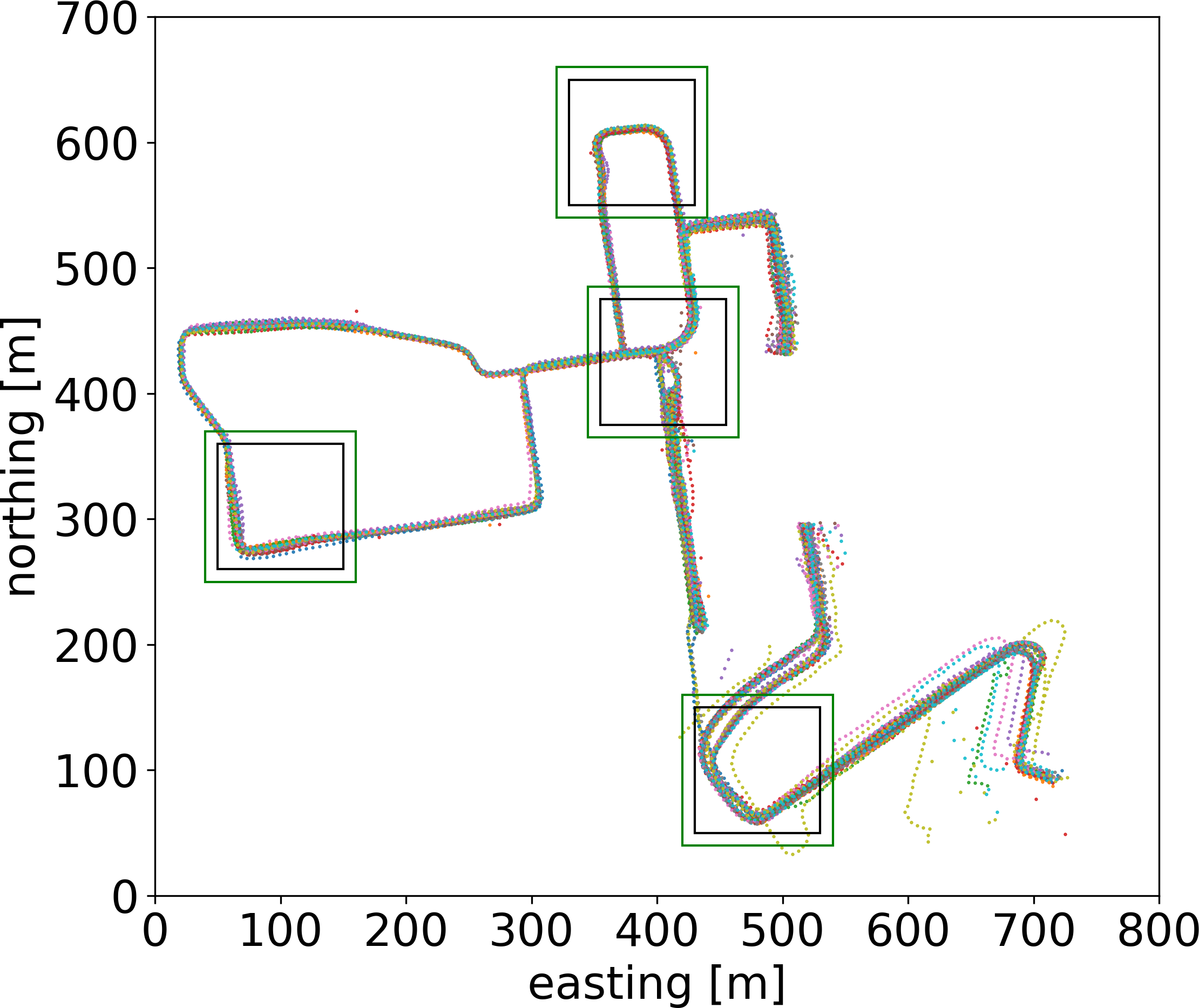}
    \caption{The visualization of 40 runs on the USyd dataset. The recorded data is divided into the testing areas (four black squares), buffer areas not used at all (between black and green squares), and the training data (outside green squares).}
    \label{fig:USyd}
\end{figure}

After skipping recordings with errors or faulty sensor readings (i.e., due to lack of GPS signal inside tunnels),
we sampled the LiDAR data with interpolated GPS so that the distance between consecutive scans from the same recording is 5 meters resulting in approx. 735 locations per each of 40 runs.
Each location for place recognition contains information from a single 3D scan that contains up to 25000 3D points with intensity measurement that might be as far as 100 m from the sensor. 
The number of points and their distribution for each location are not modified or limited compared to the raw 3D scan, i.e., the ground plane is not removed. 

The division into test, buffer and train sections is  presented in Fig.~\ref{fig:USyd}.
Randomly selected four 100$\times$100 m areas constitute test sections. 
Each test section has a 10 m buffer area that is not used for testing or training. 
The rest of the data was used to train the networks.
In total, we obtained 19138 training and 8797 test 3D point clouds, which we make publicly available in Oxford RobotCar~\cite{oxford} compatible format for future comparisons.




\subsection{Oxford RobotCar}

The Oxford RobotCar dataset~\cite{oxford} consists of data recorded over a year-long and over 1000 km, thus making it well-suited to compare 3D place recognition solutions~\cite{pointnetvlad}.
The available 3D data is generated from 2D scans accumulated over time to create a unified environment map.
The map is then divided into segments of 20 m length, with each segment containing exactly 4096 equally distributed points as opposed to a greater range and  larger number of points in 3D scans in USyd.
The 3D map segments in Oxford RobotCar do not reflect the point clouds obtained from single scans from 3D LiDARs. 
In practice, we have more points, and these points are not equally distributed coming from a single point of view.
As a result, the 3D structure of the data is not compatible with the proposed 3D spherical representation.

To enable evaluation of intensity-based solutions, we modified the typical preprocessing of the Oxford RobotCar dataset for the place recognition task introduced in~\cite{pointnetvlad} to additionally include intensity for each 3D point based on original, raw data. 
In our processing, the training and testing split is the same as in the original Oxford RobotCar. We use the same number of 4096 points per point cloud, the same length of segments, and whenever possible, we use the same preprocessing steps as for the original Oxford RobotCar dataset.
We will refer to this dataset as the Oxford RobotCar Intensity.


\subsection{KITTI}
Similarly to \cite{minklocplusplus, coral}, we decided to use the KITTI dataset (KITTI)~\cite{kitti} to test the generalization ability of place recognition while the proposed systems were trained on Oxford RobotCar or USyd datasets.
The first $170$ seconds of Sequence 00 construct the reference database. The remaining part of the sequence is used as localization queries~\cite{coral}.
The achieved performance indicates how well the place recognition system can operate in a previously unknown environment.

\subsection{Evaluation measures}

Regardless of the chosen dataset, the evaluation on the testing dataset is performed by selecting a single query location and matching it to the collection of remaining locations called database.
Matching is performed using the Euclidean distance of the descriptors computed by the network for these locations.
The performance of the place recognition system is measured with average recalls $AR@X$.
The location for $AR@X$ is assumed to be correctly recognized if the $X$ most similar locations matched from the database contain at least one location within the distance $c$ from the query location.
Among the different $X$ values, the value at 1 ($AR@1$) is the most important for robotic (i.e., SLAM) applications as it measures how often the first match from the database matches the query location.
In the SLAM scenarios, including a wrong recognition may ultimately break the localization solution.
In our evaluation, we also list the $AR@1\%$ to compare our results to the state-of-the-art solutions.

For Oxford RobotCar and KITTI datasets, we assume that the place is correctly recognized if the determined location is within $c=25~m$ of the ground truth location~\cite{pointnetvlad,minkloc3d,coral}. 
In the case of the USyd dataset, we decided to use a more challenging threshold of $c=10~m$. 
Since the LiDAR sensor is omnidirectional, the correct place recognition is determined purely by the sensor's position.


\section{Experiments}

\subsection{USyd Campus}

As a reference, we trained and tested the original version of MinkLoc3D~\cite{minkloc3d} on USyd Campus, achieving the $AR@1\%$ of $98.1\%$ and $AR@1$ of $91.7\%$.  
The evaluation of the Scan Context~\cite{scanContext} with default parameters achieved the $AR@1\%$ of $88.7\%$ and $AR@1$ of $86.0\%$ in our runs.

\begin{figure}[h!]
    \centering
    \includegraphics[width=\columnwidth]{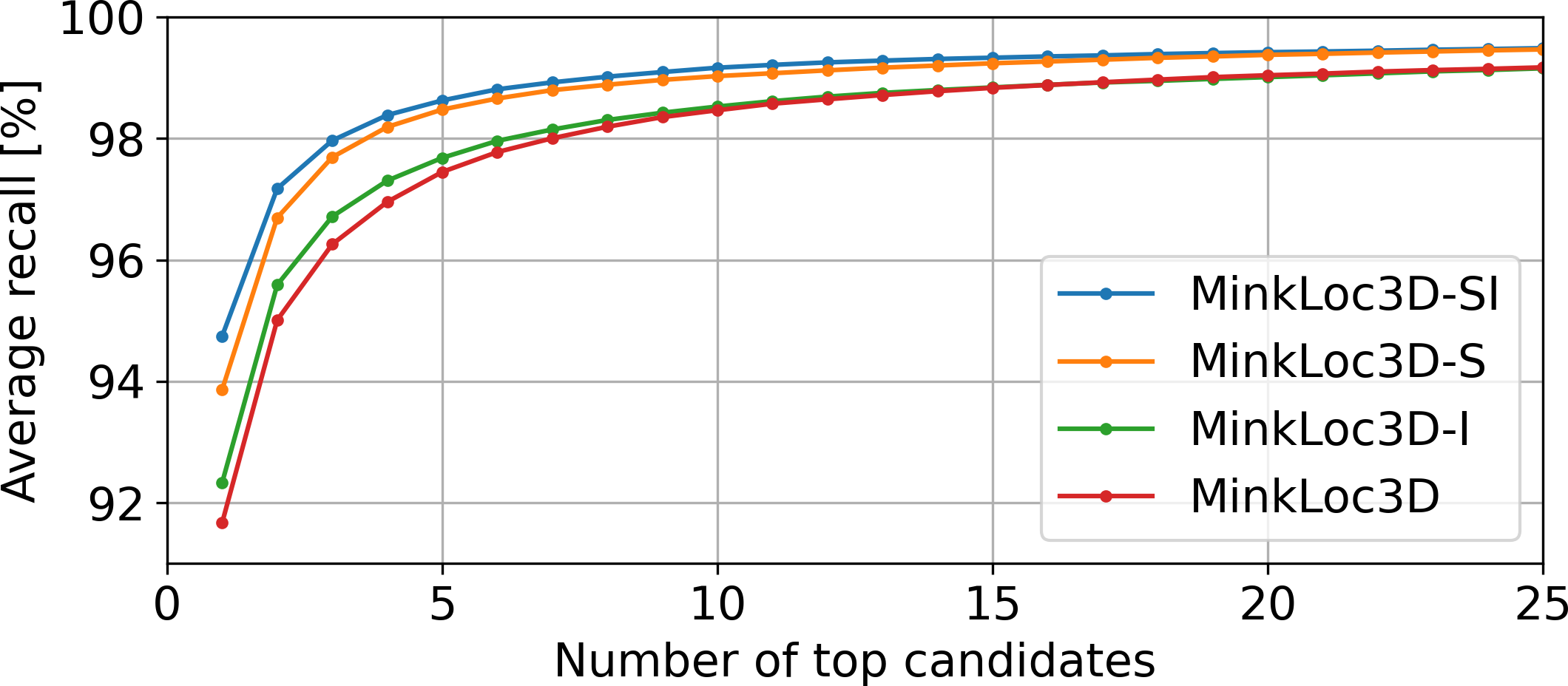}
    \caption{The visualization of $AR@X$ measure for different $X$ values for MinkLoc3D and proposed systems. MinkLoc3D-SI reports the best results, which is visible for $AR@1$. 
    Scan Context is not shown due to significantly worse results.} 
    \label{fig:resUSydAR}
\end{figure}

The obtained numerical results for proposed modifications are gathered in Tab.~\ref{tab:USyd} with a visual representation of $AR@X$ measure for X from 1 to 25 presented in Fig.~\ref{fig:resUSydAR}.

\begin{table}[htbp!]
\centering
\caption{The results obtained on the USyd dataset. Using spherical representation (MinkLoc3D-S), including intensity (MinkLoc3D-I), as well as both improvements (MinkLoc3D-SI) achieve results superior to the baseline solutions, especially for the $AR@1$ measure.}
\label{tab:USyd}
\begin{tabular}{l|c|cc}
  USyd dataset                              & Source of results & $AR@1\%$ &  $AR@1$ \\ \hline
MinkLoc3D~\cite{minkloc3d} & our evaluation & 98.1 & 91.7  \\
Scan Context~\cite{scanContext} & our evaluation &88.7 & 86.0\\
MinkLoc3D-I (our)            & our evaluation & 98.2        &   92.3         \\
MinkLoc3D-S (our)            & our evaluation & 98.8                             & 93.9                           \\
MinkLoc3D-SI (our) & our evaluation &\textbf{99.0}                                & \textbf{94.7}                          
\end{tabular}
\end{table}

MinkLoc3D-I reports an improved performance of $AR@1\%$ of $98.2\%$ and $AR@1$ of $92.3\%$,
which confirms that the intensity is complementary to the pure geometric location of the 3D points.
MinkLoc3D-S with the spherical representation of 3D point coordinates also improves the performance to $AR@1\%$ of $98.8\%$ and $AR@1\%$ of $93.9$.
The spherical representation, in this case, is better suited to capture the inter-point relations for points further from the scanner, which improves the model's ability to create high-level map features.
The most significant gains for both analyzed solutions are visible  for $AR@1$, as the $AR@1\%$ is already at a very high level and returning the single correct location is a much more challenging task.
Combining both ideas into MinkLoc3D-SI reports the best $AR@1$ (94.7) and the best $AR@1\%$ (99.0) among the evaluated solutions, as shown in Fig.~\ref{fig:resUSydAR}.


\subsection{Oxford RobotCar and Oxford RobotCar Intensity}

The results achieved when trained and evaluated on the Oxford RobotCar Intensity are gathered in Tab.~\ref{tab:oxford2}.
Adding additional intensity for MinkLoc3D-I improves the $AR@1\%$ to 98.1, which is slightly better than the version without intensity (MinkLoc3D with $AR@1\%$ of 97.6). 
The MinkLoc3D-I's $AR@1$ of 93.6 is also greater than 92.8 for MinkLoc3D. 
More significant improvements from utilizing intensity are not evident as a single 3D point in Oxford RobotCar can be measured from different distances, thus not making its intensity repeatable even for the same location.

\begin{table}[htbp!]
\centering
\caption{The comparison of results obtained on the proposed Oxford RobotCar Intensity dataset. Including intensity information improves performance (MinkLoc3D-I over MinkLoc3D) while spherical representation is not well-suited for 3D point clouds created from accumulated 2D scans (MinkLoc3D-S and MinkLoc3D-SI).}
\label{tab:oxford2}
\begin{tabular}{l|c|cc}
Oxford RobotCar Intensity      & Source of results & $AR@1\%$ & $AR@1$ \\ \hline
MinkLoc3D~\cite{minkloc3d}  & our evaluation       &  97.6            &     92.8     \\
MinkLoc3D-I (our)  & our evaluation       &  98.1            &     93.6     \\
MinkLoc3D-S (our)  & our evaluation        &  92.0             &  79.9           \\
MinkLoc3D-SI (our) & our evaluation & 93.4              &  82.2          
\end{tabular}
\end{table}

\begin{table}[htbp!]
\centering
\caption{The comparison of results obtained on the Oxford RobotCar dataset. The spherical representation is not well-suited for 3D point clouds created from accumulated 2D scans (MinkLoc3D-S), leading to poor performance compared to the state-of-the-art.}
\label{tab:oxford}
\begin{tabular}{l|c|cc}
Oxford RobotCar      & Source of results & $AR@1\%$ & $AR@1$ \\ \hline
PointNetVLAD~\cite{pointnetvlad} & \cite{pointnetvlad} & 80.3 & 63.3 \\
Scan Context~\cite{scanContext}& \cite{disco} & 81.9 & 64.6\\
PCAN~\cite{pcan} & \cite{pcan} & 83.8 & 70.7\\
DH3D-4096~\cite{dh3d} & \cite{dh3d} & 84.3 & 73.3\\
EPC-Net~\cite{epcnet} & \cite{epcnet} & 94.7 & 86.2\\
LPD-Net~\cite{lpdnet} & \cite{lpdnet} & 94.9 & 86.4\\
DISCO~\cite{disco}& \cite{disco} & 75.0 & 88.4 \\ 
SOE-Net~\cite{soenet} & \cite{soenet} & 96.4 & 89.3\\
NDT-Transformer~\cite{NDTtransformer} & \cite{NDTtransformer} & 97.7 & 93.8 \\
TransLoc3D~\cite{transloc3d} & \cite{transloc3d} & 98.5 & 95.0 \\
                        
MinkLoc3D (3D)~\cite{minkloc3d}& \cite{minkloc3d} & 97.9 & 93.8\\
MinkLoc3D-S (our)  & our evaluation        &  93.1             &  82.0           

\end{tabular}
\end{table}

MinkLoc3D-S utilizing the spherical representation of 3D points achieved the $AR@1\%$ of $92.0$ and $AR@1$ of $79.9$.
As expected, this approach harms the system's performance as points are equally distributed in the 3D point cloud. 
We also performed experiments with MinkLoc3D-SI, which improves over a version without intensity to $AR@1\%$ to 93.4 and $AR@1$ to 82.2 but works worse than MinkLoc3D-I due to spherical coordinates.

On the original formulation of the Oxford RobotCar dataset, MinkLoc3D-S performs worse than MinkLoc3D as 3D point clouds are created from multiple 2D scans as presented in Tab.~\ref{tab:oxford}. 
Similar results obtained by both methods on Oxford RobotCar and Oxford RobotCar Intensity suggest that the proposed variant of Oxford RobotCar can be used to evaluate methods when intensity information is required.



\subsection{KITTI}

We used the MinkLoc3D-SI solution trained on the Oxford RobotCar Intensity or USyd to verify its performance on the KITTI dataset as in~\cite{pointnetvlad,lpdnet,minklocplusplus}.
The only modification is the division of the quantization parameter of angle $\phi$ by four to include information from all 64 layers of the HDL-64E used in KITTI, compared to 16 layers in USyd.
The results are presented in Tab.~\ref{tab:kitti}.

\begin{table}[htbp!]
\centering
\caption{The generalization results on the KITTI dataset. MinkLoc3D-SI outperforms other solutions. Oxford RC stands for Oxford RobotCar, while Oxford RCI stands for Oxford RobotCar Intensity.}
\label{tab:kitti}
\begin{tabular}{l|cc|cc}
KITTI dataset           & \begin{tabular}[c]{@{}c@{}}Trained \\ on\end{tabular} &  \begin{tabular}[c]{@{}c@{}}Source \\ of results\end{tabular} & $AR@1\%$ &  $AR@1$ \\ \hline
PointNetVLAD~\cite{pointnetvlad} & Oxford RC & \cite{coral}  & 72.4 & -- \\
LPD-Net~\cite{lpdnet}             &  Oxford RC & \cite{coral}          & 74.6 & -- \\
MinkLoc++ (3D)~\cite{minklocplusplus} & Oxford RC &\cite{minklocplusplus}                 & 72.6 & --  \\
Scan Context~\cite{scanContext} & - & our eval. & 75.0 & 71.4 \\
MinkLoc3D~\cite{minkloc3d} & USyd & our eval.                 & 73.8 & 69.1  \\
MinkLoc3D-SI (our) & Oxford RCI & our eval. & \textbf{81.0} & 72.6 \\ 
MinkLoc3D-SI (our) & USyd & our eval. & \textbf{81.0} & \textbf{78.6} \\ 
\end{tabular}
\end{table}

In this generalization task, MinkLoc3D-SI outperforms other solutions, whether trained on the USyd or the Oxford RobotCar Intensity. 
The selected spherical representation with intensity information in MinkLoc3D-SI is a better fit for 3D LiDAR scans recorded at a single location, which is the case for the KITTI dataset. 



\section{Ablation study}


\subsection{Quantization}






Minkowski Engine performs 3D sparse convolutions on cuboids, and each cuboid can contain only one point. 
Therefore, the size of the smallest cuboid provided at the network's input determines the granularity of the input data.

\begin{figure}[h!]
    \centering
    \includegraphics[width=\columnwidth]{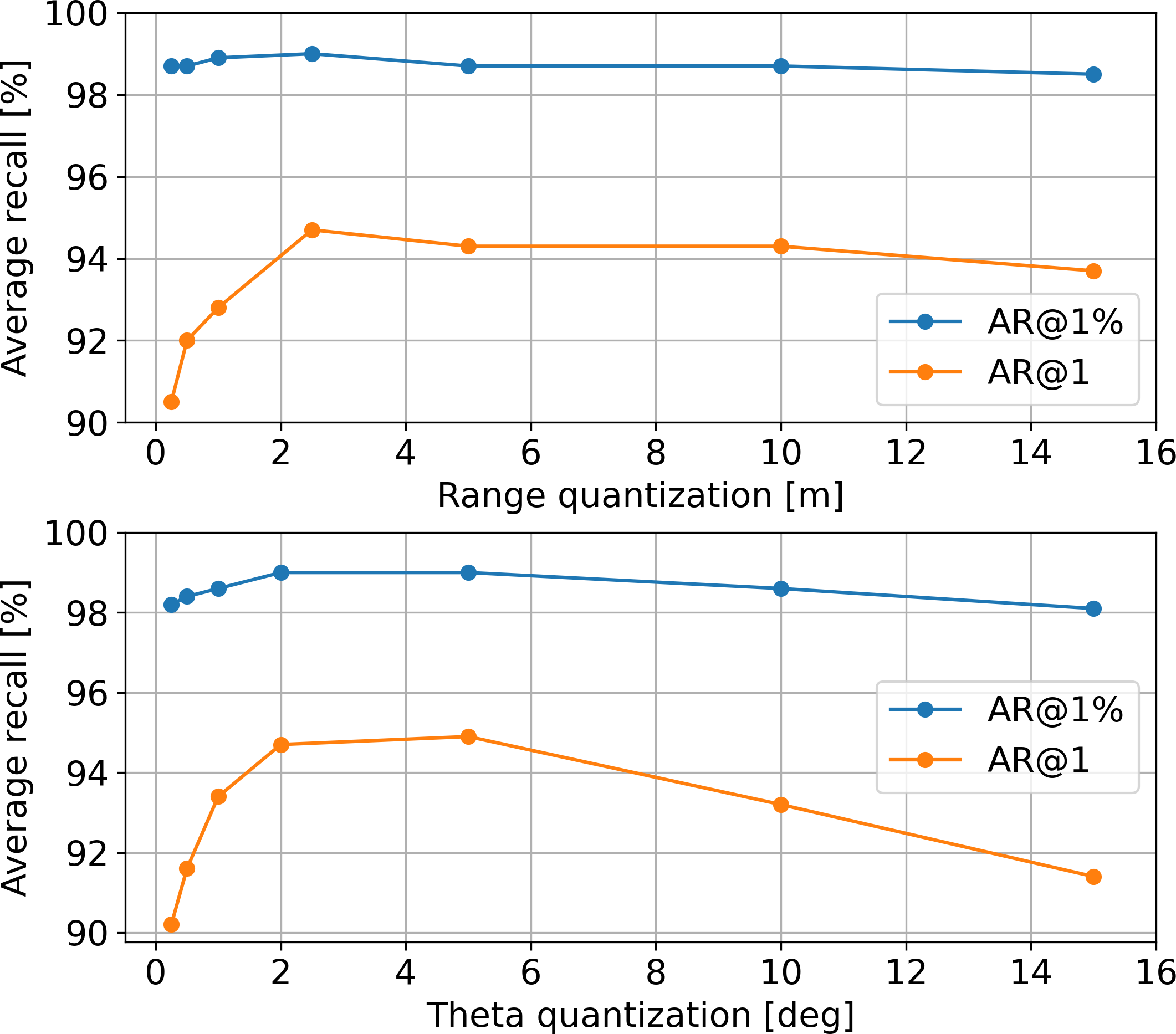}
    \caption{The influence of the range and $\theta$ quantization on the obtained results by the MinkLoc3D-SI on USyd.}
    \label{fig:quan}
    \vspace{-0.4cm}
\end{figure}

In the case of MinkLoc3D-SI, each 3D point is converted into spherical coordinates $(r, \theta, \phi)$. 
As a basis, we chose the best configuration of MinkLoc3D-SI with $r=2.5, \theta = 2\degree$.
The influence of $\phi$ was not tested as VLP-16 has only 16 scanning layers, and we assume the quantization parameter that preserves each layer.
The obtained results are presented in Fig.~\ref{fig:quan}.

The $AR@1\%$ values are similar for different quantization values, showing that the method is robust while a noticeable difference can be observed for $AR@1$.
We observe a drop in the performance for small range quantizations below 1~m as the MinkLoc3D-SI network cannot capture the high-level features correctly.
On the other hand, the granularity of the range measurement above 3 m makes the performance worse, which stems from losing necessary details in the 3D point clouds.
We determined the sweet spot for USyd to be equal to $r=2.5~m$.

A similar analysis performed for $\theta$ reveals a range of values from $\theta = 2\degree$ to $\theta = 5\degree$ resulting in a similar, best performance when it comes to $AR@1$. 
Choosing to use more cuboids ($\theta$ below $2\degree$) as well as fewer cuboids ($\theta$ above $5\degree$) results in worse $AR@1$ and $AR@1\%$ measures.

\FloatBarrier

\subsection{The maximum range of 3D LiDAR}

\begin{figure}[htbp!]
    \centering
    \includegraphics[width=\columnwidth]{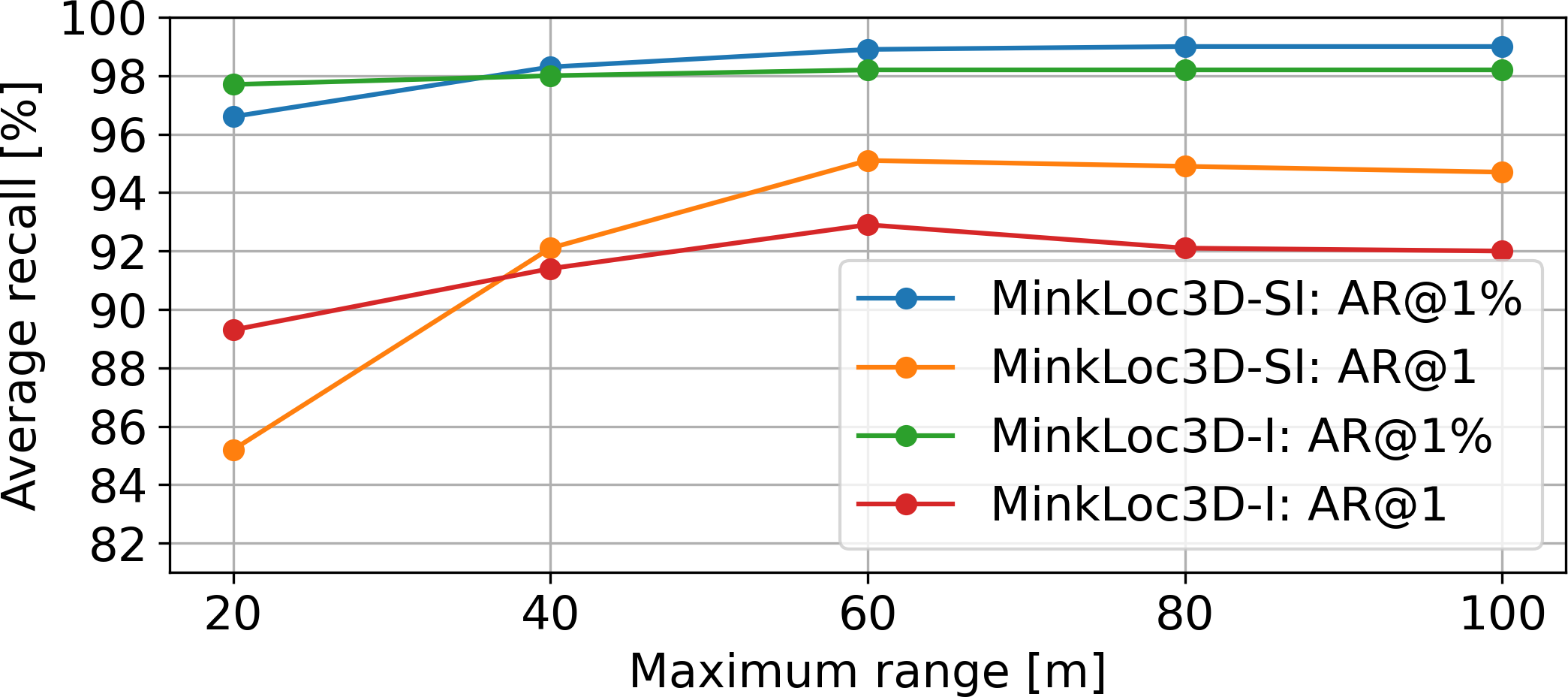}
    \caption{The maximum distance of points from the 3D LiDAR center on the results obtained by MinkLoc3D-I and MinkLoc3D-SI on USyd. A drop in performance is visible for maximum ranges below 60 m.}
    \label{fig:range}
\end{figure}

Velodyne VLP-16 used in the USyd Campus dataset has a range of approx. 100 m. MinkLoc3D-I and MinkLoc3D-SI results obtained for varying maximum range of points from the sensor are presented in Fig.~\ref{fig:range}.


The best results were obtained with the maximum range of measurement set to at least 60~m.
The 3D points are sparsely located for greater maximal ranges, making it harder to determine meaningful features for place recognition.
For these ranges, the spherical representation of MinkLoc3D-SI outperforms the Cartesian representation used in MinkLoc3D-I. 
The \textit{AR@1} for the limited ranges lower than 60 m drops while the \textit{AR@1\%} remains above $96\%$ for both methods.
For the smallest maximum range of 20~m, MinkLoc3D-I reports better \textit{AR@1} of $89.3$ than \textit{AR@1} of $85.2$ of MinkLoc3D-SI, further proving that Cartesian formulation works well when measured points are close to each other.


\FloatBarrier
\subsection{Number of points in the 3D point cloud}

\begin{figure}[htbp!]
    \centering
    \includegraphics[width=\columnwidth]{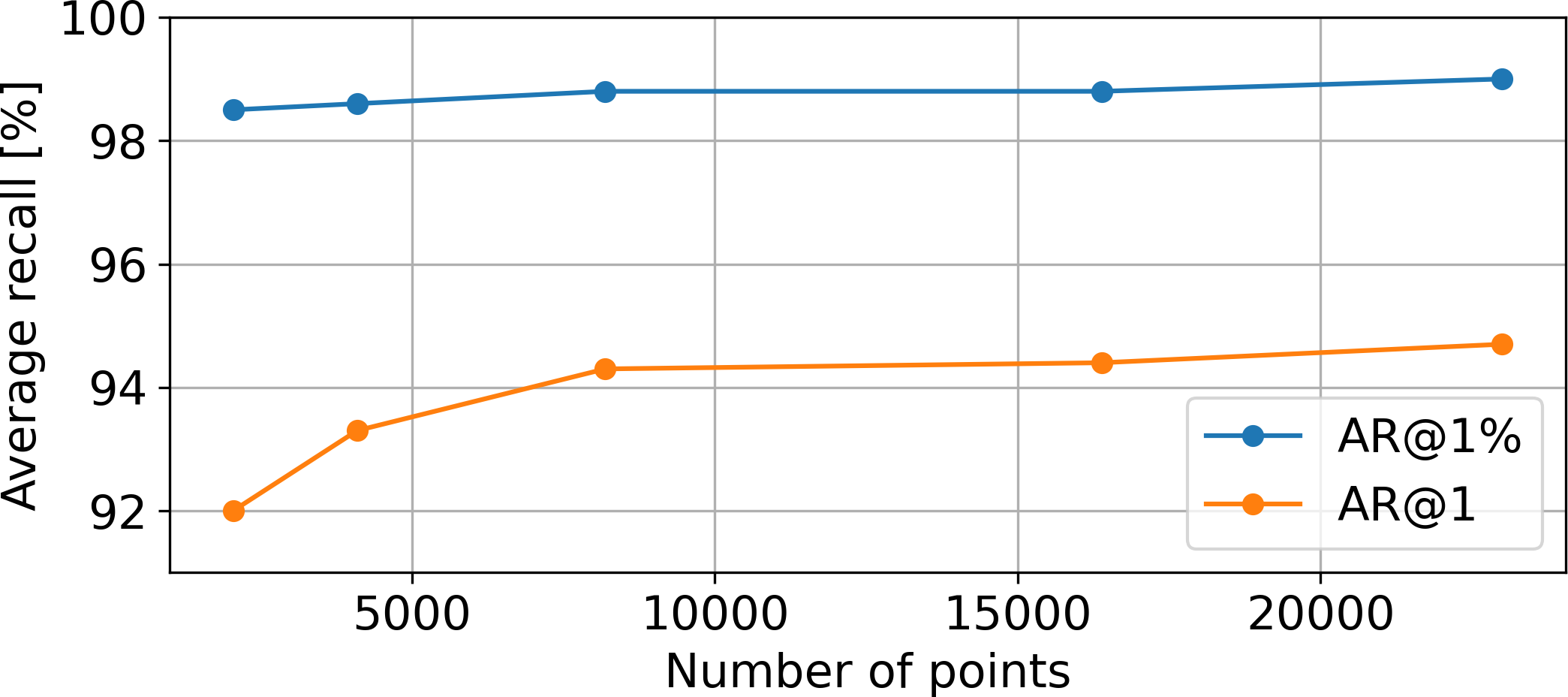}
    \caption{The influence of a randomly chosen number of points on the performance of MinkLoc3D-SI on USyd. A drop in performance is visible when the number of points is below 8192.}
    \label{fig:numberofpoints}
\end{figure}

The 3D LiDAR scans in the USyd dataset contain up to 25k points per scan, which is significantly more than 4096 points used in the Oxford RobotCar dataset. 
The performance of the proposed solutions depends on a number of randomly subsampled points, as shown in Fig.~\ref{fig:numberofpoints}.

\begin{figure}[htbp!]
    \centering
    \includegraphics[width=\columnwidth]{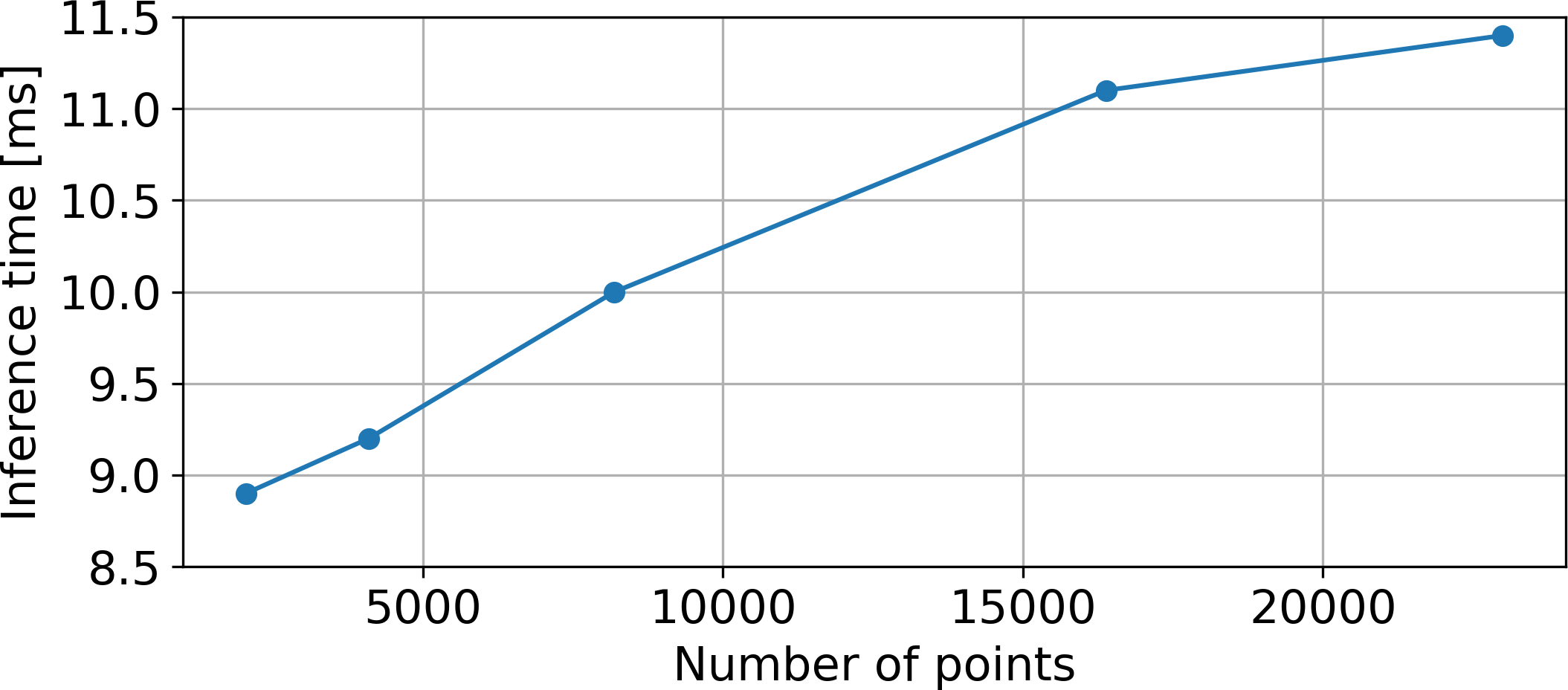}
    \caption{Inference time for a single 3D point cloud place recognition with MinkLoc3D-SI on USyd. Total inference time is below 12 ms, regardless of the number of points used.}
    \label{fig:time}
\end{figure}

The best results are obtained when all of the points from the 3D LiDAR are considered. 
This stands in contrast to the typical processing of the Oxford RobotCar dataset that limits the number of points to 4096. 
In our ablation study, the drop in performance is visible when the number of points is below 8192. 
Based on these results, we recommend using all points from the sensor for MinkLoc3D-SI as the measured inference time per 3D point cloud only increases to $11.4$~ms from $8.9$~ms on Nvidia RTX 3080 when we process 23000 points instead of 2048 as presented in Fig.~\ref{fig:time}.

\subsection{Performance across different weather conditions}

The USyd dataset was gathered along similar routes for over a year with varying different weather conditions.
Therefore, the performance of MinkLoc3D-SI depending on the weather conditions was evaluated to determine its robustness, and the results are presented in Tab.~\ref{tab:SI_weather}.

\begin{table}[htbp!]
\caption{MinkLoc3D-SI performance measured with $AR@1$ across different weather conditions: sunny (S), cloudy (C), sunny/cloudy (S/C), after/slight rain (AR), sunset (SS), very cloudy (VC). The \textbf{Imp.} stands for the percentage point improvement over MinkLoc3D in the same weather conditions.}
\label{tab:SI_weather}
\begin{tabular}{c|cccccc|c|c}
\diagbox[width=4em]{Q.}{Ref.}  & \textbf{S} & \textbf{C} & \textbf{S/C} & \textbf{AR} & \textbf{SS} & \textbf{VC} & \textbf{Mean} & \textbf{Imp.}\\ \hline
\textbf{S} & 95.1 & 93.1 & 96.0 & 95.0 & 93.4 & 96.0 & 94.6 & \cellcolor{green!25}+3.1\\
\textbf{C} & 95.2 & 92.8 & 96.0 & 94.5 & 92.7 & 94.7 & 94.5 & \cellcolor{green!25}+2.6\\
\textbf{S/C} & 96.5 & 94.7 & 96.6 & 95.6 & 94.9 & 96.4 & 95.9 & \cellcolor{green!25}+2.7\\
\textbf{AR} & 95.3 & 93.1 & 95.9 & 94.8 & 94.3 & 95.5 & 94.8 & \cellcolor{green!25}+2.8 \\
\textbf{SS} & 94.0 & 91.8 & 94.6 & 95.3 & 94.9 & 96.2 & 93.9 & \cellcolor{green!25}+3.5\\
\textbf{VC} & 95.7 & 93.0 & 95.6 & 94.7 & 94.8 & -- & 94.9 & \cellcolor{green!25}+1.7\\ \hline
\textbf{Mean} & 95.2 & 93.1 & 95.9 & 94.9 & 93.6 & 95.7 & 94.7 & \cellcolor{green!25}+2.9
\end{tabular}
\end{table}

The obtained results suggest that the performance of the proposed MinkLoc3D-SI is mostly independent of the weather conditions.
MinkLoc3D-SI yields an improvement of $AR@1$ from $1.7$ to $3.5$ percentage points over MinkLoc3D for all weather scenarios proving that spherical representation and intensity information are valuable additions in all cases.


\section{Conclusions}

In this article, we propose MinkLoc3D-SI, the sparse convolution-based method utilizing the natural, spherical representation of 3D points from a single 3D LiDAR scan, and the commonly available intensity information associated with each 3D point measurement. 
The proposed method targets the problem of place recognition when using a single scan from a 3D LiDAR.

MinkLoc3D-SI is evaluated on USyd Campus, KITTI, and Oxford RobotCar Intensity datasets.
On the USyd Campus dataset, the gains from the spherical point representation, intensity, and combined improvements are notable compared to the state-of-the-art MinkLoc3D and Scan Context.
We observe minor improvements on the proposed Oxford RobotCar Intensity dataset when intensity is used, but the spherical representation is unsuitable for map segments created from accumulated 2D scans. 
The further evaluation of the generalization ability on the KITTI dataset yields the best results among the 3D point cloud-based algorithms.
The performed ablation study confirms that the best results should be expected with rather large quantization steps and when all of the available points are processed.

The obtained results suggest that the spherical coordinates with intensity for 3D points are promising modifications to processing point clouds from a rotating 3D LiDAR and thus could be applied to other solutions with sparse 3D convolutional architecture or for other applications. 

\balance

\bibliographystyle{IEEEtran}
\bibliography{biblio}

\begin{thebibliography}{10}
\providecommand{\url}[1]{#1}
\csname url@samestyle\endcsname
\providecommand{\newblock}{\relax}
\providecommand{\bibinfo}[2]{#2}
\providecommand{\BIBentrySTDinterwordspacing}{\spaceskip=0pt\relax}
\providecommand{\BIBentryALTinterwordstretchfactor}{4}
\providecommand{\BIBentryALTinterwordspacing}{\spaceskip=\fontdimen2\font plus
\BIBentryALTinterwordstretchfactor\fontdimen3\font minus
  \fontdimen4\font\relax}
\providecommand{\BIBforeignlanguage}[2]{{%
\expandafter\ifx\csname l@#1\endcsname\relax
\typeout{** WARNING: IEEEtran.bst: No hyphenation pattern has been}%
\typeout{** loaded for the language `#1'. Using the pattern for}%
\typeout{** the default language instead.}%
\else
\language=\csname l@#1\endcsname
\fi
#2}}
\providecommand{\BIBdecl}{\relax}
\BIBdecl

\bibitem{icarcv}
K.~\.Zywanowski, A.~Banaszczyk, and M.~R. Nowicki, ``{Comparison of
  camera-based and 3D LiDAR-based place recognition across weather
  conditions},'' in \emph{2020 16th Inter. Conf. on Control, Automation,
  Robotics and Vision (ICARCV)}, 2020, pp. 886--891.

\bibitem{voxelnet}
Y.~Zhou and O.~Tuzel, ``{VoxelNet: End-to-End Learning for Point Cloud Based 3D
  Object Detection},'' in \emph{2018 IEEE/CVF Conf. on Computer Vision and
  Pattern Recognition}, 2018, pp. 4490--4499.

\bibitem{pointpillars}
A.~H. Lang, S.~Vora, H.~Caesar, L.~Zhou, J.~Yang, and O.~Beijbom,
  ``{PointPillars: Fast Encoders for Object Detection From Point Clouds},'' in
  \emph{Proc. of the IEEE/CVF Conf. on Computer Vision and Pattern Recognition
  (CVPR)}, June 2019.

\bibitem{scanContext}
G.~Kim and A.~Kim, ``Scan context: Egocentric spatial descriptor for place
  recognition within 3d point cloud map,'' in \emph{2018 IEEE/RSJ Inter. Conf.
  on Intelligent Robots and Systems (IROS)}, 2018, pp. 4802--4809.

\bibitem{scancontext++}
G.~Kim, S.~Choi, and A.~Kim, ``Scan context++: Structural place recognition
  robust to rotation and lateral variations in urban environments,'' \emph{IEEE
  Trans. on Robotics}, 2021.

\bibitem{pointnetvlad}
M.~A. Uy and G.~H. Lee, ``{PointNetVLAD: Deep point cloud based retrieval for
  large-scale place recognition},'' in \emph{Proc. of the IEEE Conf. on
  Computer Vision and Pattern Recognition}, 2018, pp. 4470--4479.

\bibitem{NDTtransformer}
Z.~Zhou, C.~Zhao, D.~Adolfsson, S.~Su, Y.~Gao, T.~Duckett, and L.~Sun,
  ``Ndt-transformer: Large-scale 3d point cloud localisation using the normal
  distribution transform representation,'' in \emph{2021 IEEE International
  Conference on Robotics and Automation (ICRA)}, 2021, pp. 5654--5660.

\bibitem{minkloc3d}
J.~Komorowski, ``{MinkLoc3D: Point Cloud Based Large-Scale Place
  Recognition},'' in \emph{Proc. of the IEEE/CVF Winter Conf. on Applications
  of Computer Vision (WACV)}, 2021, pp. 1790--1799.

\bibitem{minklocplusplus}
J.~Komorowski, M.~Wysoczanska, and T.~Trzcinski, ``{MinkLoc++: Lidar and
  Monocular Image Fusion for Place Recognition},'' in \emph{2021 Inter. Joint
  Conf. on Neural Networks (IJCNN)}, 2021, pp. 1--8.

\bibitem{attdlnet}
T.~Barros, L.~Garrote, R.~Pereira, C.~Premebida, and U.~J. Nunes, ``{AttDLNet:
  Attention-based DL Network for 3D LiDAR Place Recognition},'' \emph{arXiv
  preprint arXiv:2106.09637}, 2021.

\bibitem{transloc3d}
T.-X. Xu, Y.-C. Guo, Y.-K. Lai, and S.-H. Zhang, ``{TransLoc3D: Point Cloud
  based Large-scale Place Recognition using Adaptive Receptive Fields},''
  \emph{arXiv preprint arXiv:2105.11605}, 2021.

\bibitem{pcan}
W.~Zhang and C.~Xiao, ``{PCAN: 3D Attention Map Learning Using Contextual
  Information for Point Cloud Based Retrieval},'' in \emph{Proc. of the
  IEEE/CVF Conf. on Computer Vision and Pattern Recognition (CVPR)}, June 2019.

\bibitem{lpdnet}
Z.~Liu, S.~Zhou, C.~Suo, P.~Yin, W.~Chen, H.~Wang, H.~Li, and Y.-H. Liu,
  ``{LPD-Net: 3D Point Cloud Learning for Large-Scale Place Recognition and
  Environment Analysis},'' in \emph{Proc. of the IEEE Inter. Conf. on Computer
  Vision}, 2019, pp. 2831--2840.

\bibitem{dh3d}
J.~Du, R.~Wang, and D.~Cremers, ``Dh3d: Deep hierarchical 3d descriptors for
  robust large-scale 6dof relocalization,'' in \emph{European Conf. on Computer
  Vision}.\hskip 1em plus 0.5em minus 0.4em\relax Springer, 2020, pp. 744--762.

\bibitem{dagc}
Q.~Sun, H.~Liu, J.~He, Z.~Fan, and X.~Du, ``{DAGC: Employing Dual Attention and
  Graph Convolution for Point Cloud Based Place Recognition},'' in \emph{Proc.
  of the 2020 Inter. Conf. on Multimedia Retrieval}, ser. ICMR '20.\hskip 1em
  plus 0.5em minus 0.4em\relax New York, NY, USA: Association for Computing
  Machinery, 2020, pp. 224--232.

\bibitem{epcnet}
L.~Hui, M.~Cheng, J.~Xie, and J.~Yang, ``{Efficient 3D Point Cloud Feature
  Learning for Large-Scale Place Recognition},'' \emph{arXiv preprint
  arXiv:2101.02374}, 2021.

\bibitem{soenet}
Y.~Xia, Y.~Xu, S.~Li, R.~Wang, J.~Du, D.~Cremers, and U.~Stilla, ``Soe-net: A
  self-attention and orientation encoding network for point cloud based place
  recognition,'' in \emph{Proceedings of the IEEE/CVF Conference on Computer
  Vision and Pattern Recognition (CVPR)}, June 2021, pp. 11\,348--11\,357.

\bibitem{usyd}
\BIBentryALTinterwordspacing
W.~Zhou and et~al., ``{The USyd Campus Dataset},'' 2019. [Online]. Available:
  \url{https://dx.doi.org/10.21227/sk74-7419}
\BIBentrySTDinterwordspacing

\bibitem{oxford}
W.~Maddern, G.~Pascoe, C.~Linegar, and P.~Newman, ``{1 Year, 1000km: The Oxford
  RobotCar Dataset},'' \emph{The Inter. Journal of Robotics Research (IJRR)},
  vol.~36, no.~1, pp. 3--15, 2017.

\bibitem{kitti}
A.~Geiger, P.~Lenz, C.~Stiller, and R.~Urtasun, ``{Vision Meets Robotics: The
  KITTI Dataset},'' \emph{Inter. Journal of Robotic Research (IJRR)}, vol.~32,
  no.~11, pp. 1231--1237, Sep. 2013.

\bibitem{pointnet}
R.~Q. Charles, H.~Su, M.~Kaichun, and L.~J. Guibas, ``{PointNet: Deep Learning
  on Point Sets for 3D Classification and Segmentation},'' in \emph{2017 IEEE
  Conf. on Computer Vision and Pattern Recognition (CVPR)}, 2017, pp. 77--85.

\bibitem{pointnet++}
C.~R. Qi, L.~Yi, H.~Su, and L.~J. Guibas, ``{PointNet++: Deep Hierarchical
  Feature Learning on Point Sets in a Metric Space},'' in \emph{Proc. of the
  31st Inter. Conf. on Neural Information Processing Systems}, ser.
  NIPS'17.\hskip 1em plus 0.5em minus 0.4em\relax Curran Associates Inc., 2017,
  pp. 5105--5114.

\bibitem{minkowski}
C.~Choy, J.~Gwak, and S.~Savarese, ``{4D Spatio-Temporal ConvNets: Minkowski
  Convolutional Neural Networks},'' in \emph{{Proc. of the IEEE Conf. on
  Computer Vision and Pattern Recognition}}, 2019, pp. 3075--3084.

\bibitem{sparse2}
E.~Elsen, M.~Dukhan, T.~Gale, and K.~Simonyan, ``Fast sparse convnets,'' in
  \emph{Proceedings of the IEEE/CVF Conference on Computer Vision and Pattern
  Recognition (CVPR)}, June 2020.

\bibitem{spvnet}
H.~Tang, Z.~Liu, S.~Zhao, Y.~Lin, J.~Lin, H.~Wang, and S.~Han, ``{Searching
  Efficient 3D Architectures with Sparse Point-Voxel Convolution},'' in
  \emph{European Conf. on Computer Vision}, 2020.

\bibitem{overlapnet}
X.~Chen, T.~L\"abe, A.~Milioto, T.~R\"ohling, O.~Vysotska, A.~Haag, J.~Behley,
  and C.~Stachniss, ``{OverlapNet: Loop Closing for LiDAR-based SLAM},'' in
  \emph{Proc. of Robotics: Science and Systems (RSS)}, 2020.

\bibitem{netvlad}
R.~Arandjelovi\'{c}, P.~Gronat, A.~Torii, T.~Pajdla, and J.~Sivic, ``{NetVLAD:
  CNN Architecture for Weakly Supervised Place Recognition},'' \emph{IEEE
  Trans. on Pattern Analysis and Machine Intelligence}, vol.~40, no.~6, pp.
  1437--1451, 2018.

\bibitem{segmap}
R.~Dubé, A.~Cramariuc, D.~Dugas, H.~Sommer, M.~Dymczyk, J.~Nieto, R.~Siegwart,
  and C.~Cadena, ``{SegMap: Segment-based mapping and localization using
  data-driven descriptors},'' \emph{The Inter. Journal of Robotics Research},
  vol.~39, no. 2-3, pp. 339--355, 2020.

\bibitem{janps}
J.~Wietrzykowski and P.~Skrzypczy\'nski, ``{On the descriptive power of LiDAR
  intensity images for segment-based loop closing in 3-D SLAM},'' in \emph{2021
  IEEE/RSJ Inter. Conf. on Intelligent Robots and Systems (IROS)}, 2021, pp.
  64--70.

\bibitem{locus}
K.~Vidanapathirana, P.~Moghadam, B.~Harwood, M.~Zhao, S.~Sridharan, and
  C.~Fookes, ``Locus: Lidar-based place recognition using spatiotemporal
  higher-order pooling,'' in \emph{2021 IEEE International Conference on
  Robotics and Automation (ICRA)}, 2021, pp. 5075--5081.

\bibitem{disco}
X.~Xu, H.~Yin, Z.~Chen, Y.~Li, Y.~Wang, and R.~Xiong, ``{DiSCO: Differentiable
  Scan Context With Orientation},'' \emph{IEEE Robotics and Automation
  Letters}, vol.~6, no.~2, pp. 2791--2798, 2021.

\bibitem{intensitySC}
H.~Wang, C.~Wang, and L.~Xie, ``Intensity scan context: Coding intensity and
  geometry relations for loop closure detection,'' in \emph{2020 IEEE
  International Conference on Robotics and Automation (ICRA)}, 2020, pp.
  2095--2101.

\bibitem{li2021ssc}
L.~{Li}, X.~{Kong}, X.~{Zhao}, T.~{Huang}, W.~{Li}, F.~{Wen}, H.~{Zhang}, and
  Y.~{Liu}, ``Ssc: Semantic scan context for large-scale place recognition,''
  in \emph{2021 IEEE/RSJ Inter. Conf. on Intelligent Robots and Systems
  (IROS)}, 2021.

\bibitem{bvmatch}
L.~Luo, S.-Y. Cao, B.~Han, H.-L. Shen, and J.~Li, ``{BVMatch: Lidar-Based Place
  Recognition Using Bird's-Eye View Images},'' \emph{IEEE Robotics and
  Automation Letters}, vol.~6, no.~3, pp. 6076--6083, 2021.

\bibitem{ecanet}
Q.~Wang, B.~Wu, P.~Zhu, P.~Li, W.~Zuo, and Q.~Hu, ``{ECA-Net: Efficient Channel
  Attention for Deep Convolutional Neural Networks},'' in \emph{Proc. of the
  IEEE/CVF Conf. on Computer Vision and Pattern Recognition (CVPR)}, June 2020.

\bibitem{gem}
F.~Radenović, G.~Tolias, and O.~Chum, ``{Fine-Tuning CNN Image Retrieval with
  No Human Annotation},'' \emph{IEEE Trans. on Pattern Analysis and Machine
  Intelligence}, vol.~41, no.~7, pp. 1655--1668, 2019.

\bibitem{fpn}
T.-Y. Lin, P.~Dollár, R.~Girshick, K.~He, B.~Hariharan, and S.~Belongie,
  ``Feature pyramid networks for object detection,'' in \emph{2017 IEEE Conf.
  on Computer Vision and Pattern Recognition (CVPR)}, 2017, pp. 936--944.

\bibitem{triplet2}
V.~Lepetit, ``\BIBforeignlanguage{English}{Learning descriptors for object
  recognition and 3d pose estimation},'' in
  \emph{\BIBforeignlanguage{English}{Conf. on Computer Vision and Pattern
  Recognition}}, 2015, pp. 1--10.

\bibitem{triplet}
A.~Hermans, L.~Beyer, and B.~Leibe, ``{In Defense of the Triplet Loss for
  Person Re-Identification},'' \emph{arXiv preprint arXiv:1703.07737}, 2017.

\bibitem{lidarintensity}
J.~Guo, P.~V.~K. Borges, C.~Park, and A.~Gawel, ``{Local Descriptor for Robust
  Place Recognition Using LiDAR Intensity},'' \emph{IEEE Robotics and
  Automation Letters}, vol.~4, no.~2, pp. 1470--1477, 2019.

\bibitem{lidarintensity2}
I.~A. Barsan, S.~Wang, A.~Pokrovsky, and R.~Urtasun, ``Learning to localize
  using a lidar intensity map,'' in \emph{2nd Conference on Robot Learning
  (PMLR)}, 2020, pp. 605--616.

\bibitem{coral}
Y.~Pan, X.~Xu, W.~Li, Y.~Wang, and R.~Xiong, ``{Coral: Colored structural
  representation for bi-modal place recognition},'' \emph{arXiv preprint
  arXiv:2011.10934}, 2020.

\end{thebibliography}
\end{document}